\def\year{2019}\relax
\documentclass[letterpaper]{article} 
\usepackage{aaai19}  

\usepackage{url}
\usepackage{latexsym}
\usepackage{subcaption}
\usepackage{graphicx}
\usepackage{amsmath}
\usepackage{color}
\usepackage{tabularx}
\usepackage{multirow}
\usepackage{url}
\usepackage{amsmath}

\usepackage{times}  
\usepackage{helvet}  
\usepackage{courier}  
\usepackage{url}  
\usepackage{graphicx}  
\usepackage{makecell}
\usepackage{comment}
\begin{document}
%
\title{AutoSense Model for Word Sense Induction}

\author{
    Reinald Kim Amplayo \quad Seung-won Hwang \quad Min Song\\
    Yonsei University \\
    {\tt \{rktamplayo, seungwonh, min.song\}@yonsei.ac.kr}
}

\maketitle
\begin{abstract}
 Word sense induction (WSI), or the task of automatically discovering multiple senses or meanings of a word, has three main challenges: domain adaptability, novel sense detection, and sense granularity flexibility. While current latent variable models are known to solve the first two challenges, they are not flexible to different word sense granularities, which differ very much among words, from \textit{aardvark} with one sense, to \textit{play} with over 50 senses. Current models either require hyperparameter tuning or nonparametric induction of the number of senses, which we find both to be ineffective. Thus, we aim to eliminate these requirements and solve the \textbf{sense granularity problem} by proposing \textbf{AutoSense}, a latent variable model based on two observations: (1) senses are represented as a distribution over topics, and (2) senses generate pairings between the target word and its neighboring word. These observations alleviate the problem by (a) throwing garbage senses and (b) additionally inducing fine-grained word senses. Results show great improvements over the state-of-the-art models on popular WSI datasets. We also show that AutoSense is able to learn the appropriate sense granularity of a word. Finally, we apply AutoSense to the unsupervised author name disambiguation task where the sense granularity problem is more evident and show that AutoSense is evidently better than competing models. We share our data and code here: \url{https://github.com/rktamplayo/AutoSense}.
\end{abstract}

\setlength{\abovedisplayskip}{0pt}%
\setlength{\belowdisplayskip}{0pt}%
\setlength{\abovedisplayshortskip}{0pt}%
\setlength{\belowdisplayshortskip}{0pt}%
\setlength{\jot}{0pt}

\section{Introduction}

\begin{figure}
    \centering
    \includegraphics[width=0.45\textwidth]{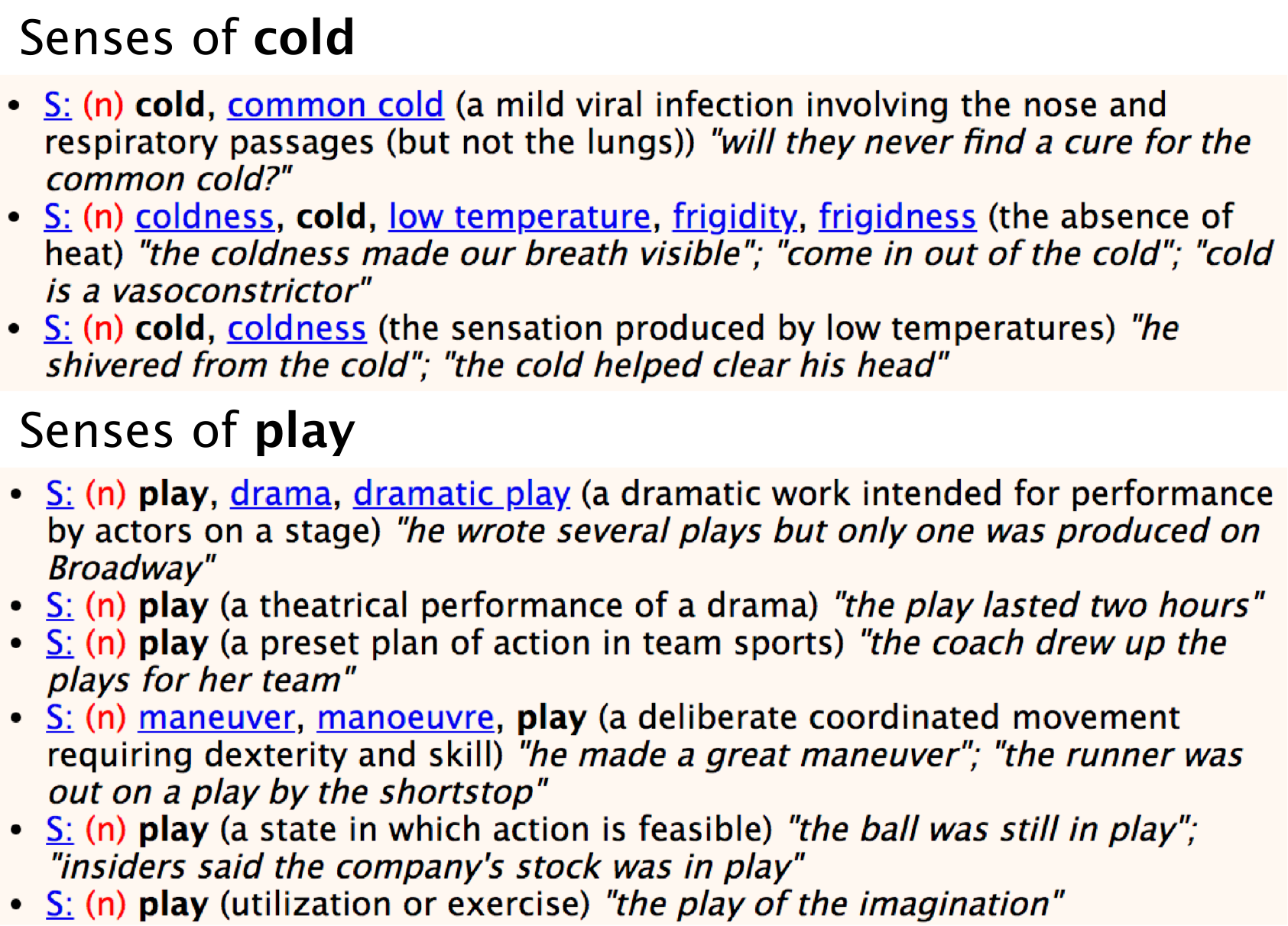}
    \caption{Three senses of the noun \textit{cold} and six of 17 senses of the noun \textit{play} in WordNet. Sense granularity problem refers to the inflexibility of the model to the different number of senses different words may have (i.e. 3 vs. 17).}
    \label{fig:wsi_example}
\end{figure}

Word sense induction (WSI) is the task where given an ambiguous target word (e.g. \textit{cold}) and texts where the word is used, we automatically discover its multiple senses or meanings (e.g. (1) \textit{nose infection}, (2) \textit{absence of heat}, etc.). We show examples of words with multiple senses and example usage in a text\footnote{All sense meanings are copied from WordNet: \url{http://wordnetweb.princeton.edu/perl/webwn}} in Figure \ref{fig:wsi_example}. It is distinct from its similar supervised counterpart, word sense disambiguation (WSD) \cite{stevenson2003word}, because WSI models should consider the following challenges due to its unsupervised nature: (C1) adaptability to new domains, (C2) ability to detect novel senses, and (C3) flexibility to different word sense granularities \cite{jurgens2013semeval}. Another task similar to the WSI is the unsupervised author name disambiguation (UAND) task \cite{song2007efficient}, where it aims to automatically find different authors, instead of words, with the same name.

In this paper, we consider a latent variable modeling approach to WSI problem as it is proven to be more effective than other approaches \cite{chang2014inducing,komninos2016structured}. Specifically, we look into methods based on Latent Dirichlet Allocation (LDA) \cite{blei2003latent}, a topic modeling method that automatically discovers the topics underlying a set of documents using Dirichlet priors to infer the multinomial distribution over words and topics. LDA naturally answers two of the three main problems mentioned above, i.e. (C1) and (C2), of the WSI task \cite{brody2009bayesian}. However, it is not flexible with regards to (C3), or the \textbf{sense granularity problem}, as it requires the users to specify the number of senses: Current systems \cite{wang2015sense,chang2014inducing} required to set the number of senses to a small number (set to 3 or 5 in the literature) to get a good accuracy, however many words may have a large number of senses, e.g. \textit{play} in Figure \ref{fig:wsi_example}.

\begin{figure}
	\centering
	\includegraphics[width=0.45\textwidth]{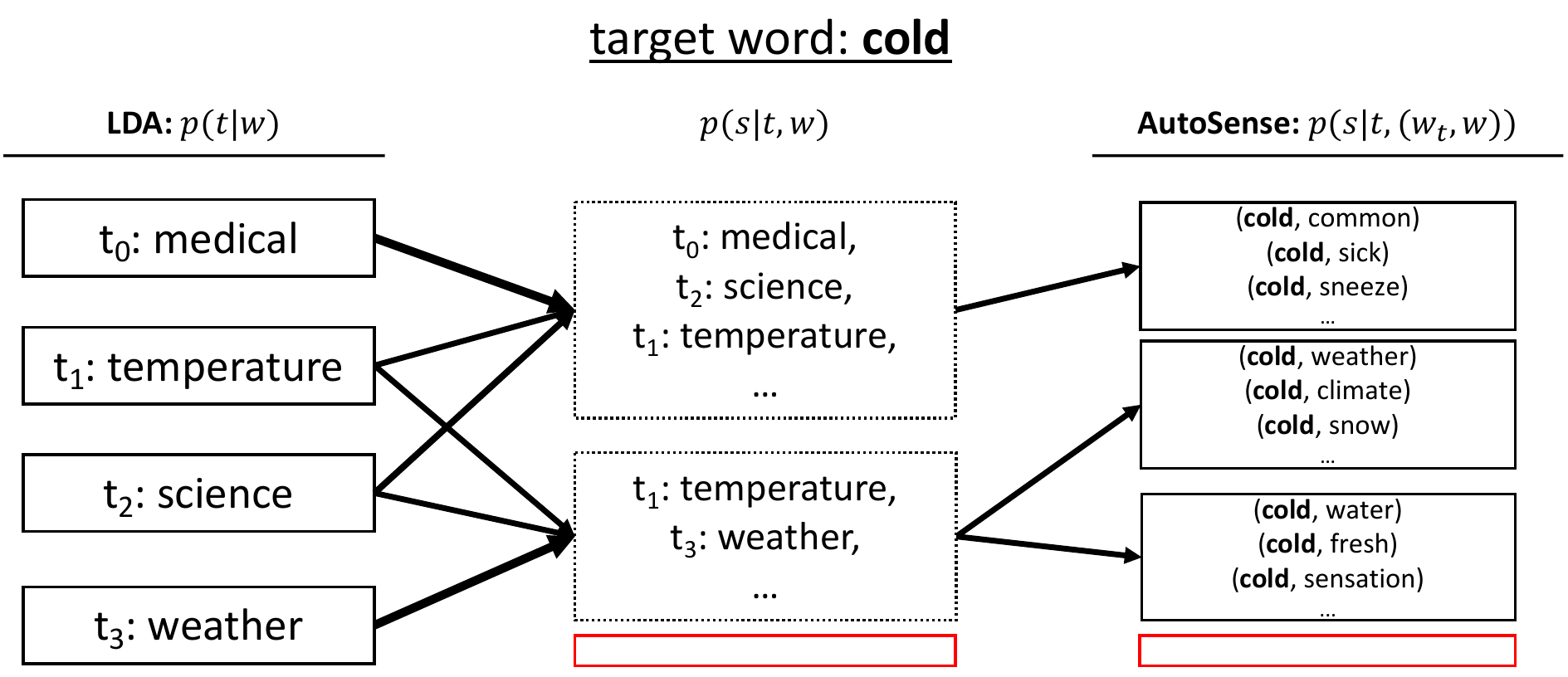}
	\caption{Example induced senses when the target word is \textit{cold} from LDA and AutoSense. Applying our observations to LDA introduces both garbage and fine-grained senses.}
	\label{fig:intuition}
\end{figure}

To this end, we propose a latent variable model called \textbf{AutoSense} that solves all the challenges of WSI, including overcoming the sense granularity problem. Consider Figure \ref{fig:intuition} on finding the senses of the target word \textit{cold}. An LDA model naively considers the topics as senses and thus differentiates the usage of \textit{cold} in the \textit{medical} and \textit{science} domains, even though the same sense is commonly used in the two domains. This results in too many senses induced by the model. We extend LDA using two observations. First, we introduce a separate latent variable for senses, which can be represented as a distribution over topics. This introduces more accurate induced senses (e.g. the \textit{cold: nose infection} sense can be from a mixture of medical, science, and temperature topics), as well as \textbf{garbage senses} (colored \textcolor{red}{red} in the figure) as most topic distributions will not be assigned to any instance. Second, we enforce senses to generate target-neighbor pairs, a pair $(w_t, w)$ which consists of the target word $w_t$ and one of its neighboring word $w$, at once. This separates the topic distributions into \textbf{fine-grained senses} based on lexical semantic features easily captured by the target-neighbor pairs. For example, the \textit{cold: absence of heat} and the \textit{cold: sensation from low temperature} senses are both related to temperature, but have different syntactic and semantic usage.

By applying the two observations above, AutoSense removes the strict requirement on correctly setting the number of senses by throwing garbage senses and introducing fine-grained senses. Nonparametric models \cite{teh2004sharing,lau2013unimelb} have also been used to solve this problem by automatically inducing the number of senses, however our experiments show that these models are less effective than parametric models and induce incorrect number of senses. Our proposed model is parametric, and is also able to adapt to the different number of senses of different words, even when the number of senses is set to an arbitrarily large number. Moreover, the model can also be used in other tasks such as UAND where the variance in the number of senses is large. To the best of our knowledge, we are the first to experiment extensively on the sense granularity problem of parametric latent variable models.

In our experiments, we estimate the parameters of the model using collapsed Gibbs sampling and get the sense distribution of each instance as the WSI solution. We evaluate our model using the SemEval 2010 and 2013 WSI datasets \cite{manandhar2010semeval,jurgens2013semeval}. Results show that AutoSense performs superior than previous state-of-the-art models. We also provide analyses and experiments that shows how AutoSense overcomes the issue on sense granularity. Finally, we show that our model performs the best on unsupervised author name disambiguation (UAND), where the sense granularities are extremely varied.

\section{Related Work}

Previous works on WSI used context vectors and attributes \cite{almuhareb2006msda}, pretrained classification systems \cite{tsvetkov2014augmenting}, and alignment of parallel corpus \cite{yao2012expectations}. In the most recent shared task on WSI \cite{jurgens2013semeval}, top models used lexical substitution method (\textbf{AI-KU}) \cite{baskaya2013ai} and Hierarchical Dirichlet Process trained with additional instances (\textbf{Unimelb}) \cite{lau2013unimelb}.

Latent variable models such as LDA \cite{blei2003latent} are used to induce the word sense of a target word after rigorous preprocessing and feature extraction (\textbf{LDA}, \textbf{Spectral}) \cite{goyal2014unsupervised}.  More recent models introduced a latent variable for the sense of a word, with the assumption that a sense has multiple concepts (\textbf{HC}, \textbf{HC+Zipf}) \cite{chang2014inducing} and that topics and senses should be inferred jointly (\textbf{STM}) \cite{wang2015sense}. In this paper, we also use a separate sense latent variable, however we show boost in performance by representing it with more versatility and by incorporating the use of target-neighbor pairs. HC was also extended to a nonparametric model (\textbf{BNP-HC}) \cite{teh2004sharing} in order to automatically set the number of senses of a word, providing flexibility to the sense granularity \cite{yao2011nonparametric,lau2012word,lau2013unimelb}. In our experiments, we show that the sense granularity induced from nonparametric models are incorrect making the models less effective.

Recent inclusions to the WSI models are neural-based dense distributional representation models. STM also used word embeddings \cite{mikolov2013distributed} to assign similarity weights during inference (\textbf{STM+w2v}) \cite{wang2015sense}. Existing sense embeddings are also used to perform word sense induction (\textbf{CRP-PPMI}, \textbf{SE-WSI-fix}, \textbf{WG}, \textbf{DIVE}) \cite{song2016word,pelevina2016making,chang2018efficient}. These models, on their own, do not perform well on the WSI task until recently when embeddings of words and their dependencies are used to construct a probabilistic model (\textbf{MCC}) \cite{komninos2016structured}. We show that neural-based embeddings are still ineffective for this task and that our model performs better than these models as well.

In the unsupervised author name disambiguation (UAND) domain, LDA-based models have also been used \cite{shu2009latent} to employ text features for the task, while non-text features such as co-authors, publication venue, year, and citations are found to be stronger features \cite{tang2012unified}. In this paper, we study on how to improve the performance of text features for UAND using latent variable models, which can later be combined with non-text features in the future work.

\section{Proposed Model}
\begin{figure}[t]
    \centering
    \includegraphics[width=0.4\textwidth]{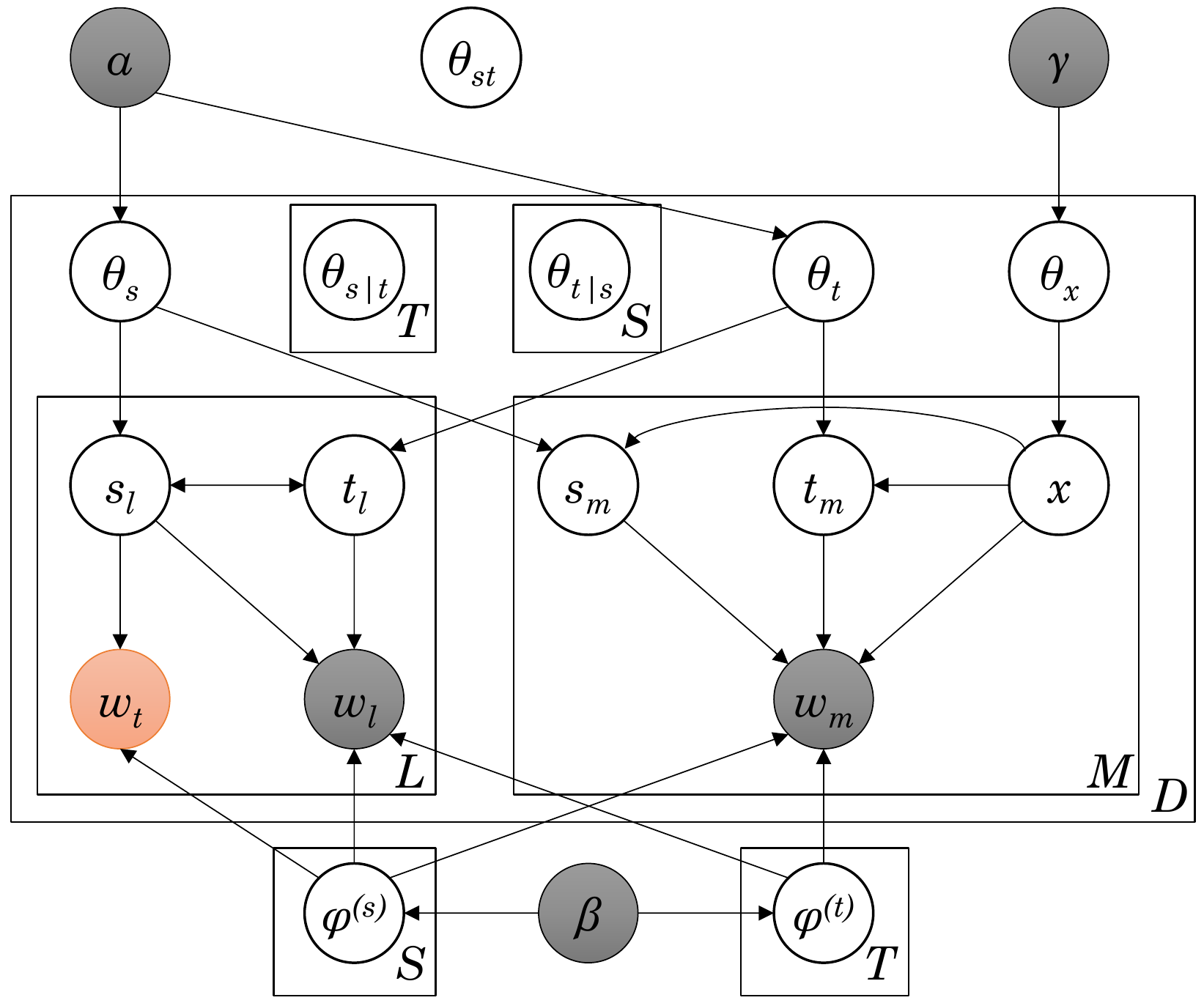}
    \caption{Graphical representation of AutoSense. Nodes are random variables, edges are dependencies, and plates are replications. Nodes shaded in black are observed. The node shaded in red is the observed target word. The dependency edges of $\theta_{s|t}$, $\theta_{t|s}$, and $\theta_{st}$ are not shown for clarity: They are all generated by the Dirichlet prior $\alpha$. Moreover, sense variables are dependent to $\theta_{s|t}$ and $\theta_{st}$, while topic variables are dependent to $\theta_{t|s}$ and $\theta_{st}$.}
    \label{fig:condsptm}
\end{figure}
\begin{table}[t]
    \scriptsize
    \centering
    \begin{tabularx}{0.47\textwidth}{|cX|}
        \hline
        $D$ & \# of documents \\
        $L$ & \# of local context words \\
        $M$ & \# of global context words \\
        $S$ & \# of senses \\
        $T$ & \# of topics \\
        $V$ & vocabulary size \\
        $w_t$ & target word \\
        $w_l,w_m$ & word in local/global context \\
        $s_l,s_m$ & sense in local/global context \\
        $t_l,t_m$ & topic in local/global context \\
        $x$ & sense/topic switch \\
        $\theta_s,\theta_t,\theta_{x}$ & multinomial distribution over senses/topics/switches \\
        $\theta_{s|t},\theta_{t|s}$ & multinomial distribution over senses/topics given topics/senses \\
        $\theta_{st}$ & multinomial distribution over sense \& topic pairs \\
        $\phi^{(s)},\phi^{(t)}$ & multinomial distribution over words \\
        $\alpha$ & Dirichlet prior over $\theta$s, except $\theta_{x}$ \\
        $\beta$ & Dirichlet prior over $\phi$s \\
        $\gamma$ & Dirichlet prior over $\theta_{x}$ \\
        \hline
    \end{tabularx}
    \caption{Meanings of the notations in AutoSense}
    \label{tab:notations}
\end{table}

There are two reasons why Latent Dirichlet Allocation (LDA) \cite{blei2003latent} is not effective for WSI. First, LDA tries to give instance assignments to all senses even when it is unnecessary. For example, when the number of senses $S$ is set to 10, the model tries to assign all the senses to all instances even when the original number of senses of a target word is 3. LDA extensions \cite{wang2015sense,chang2014inducing} mitigated this problem by setting $S$ to a small number (e.g. 3 or 5). However, this is not a good solution because there are many words with more than five senses. Second, LDA and its extensions do not consider the existence of fine-grained senses. For example, the \textit{cold: absence of heat} and the \textit{cold: sensation from low temperature} senses are fine-grained senses because they are similarly related to temperature yet have different usage. 

\subsection{AutoSense Model}

To solve the problems above, we propose to extend LDA in two parts. First, we introduce a new latent variable, apart from the topic latent variable, to represent word senses. Previous works also attempted to introduce a separate sense latent variable to generate all the words \cite{chang2014inducing}, or to generate only the neighboring words within a local context, decided by a strict user-specified window \cite{wang2015sense}. We improve by softening the strict local context assumption by introducing a switch variable which decides whether a word not in a local context should be generated by conditioning also on the sense latent variable. Our experiments show that our sense representation provides superior improvements from previous models.

Second, we force the model to generate target-neighbor pairs at once in the local context, instead of generating words one by one. A target-neighbor pair $(w_t, w)$ consists of the target word $w_t$ and a neighboring word $w$ in the local context. For example, the target-neighbor pairs in ``\textit{cold} snowy weather'', where $w_t$ is \textit{cold}, are $(cold,snowy)$ and $(cold,weather)$. These pairs give explicit information on the lexical semantics of the target word given the neighboring words. In our running example (Figure \ref{fig:intuition}), the \textit{cold: absence of heat} and the \textit{cold: sensation from low temperature} senses can be easily differentiated when we are given the target-neighbor pairs $(cold,weather)$ and $(cold,climate)$ for the former, and $(cold,water)$ and $(cold,fresh)$ for the latter sense, rather than the individual words.

These extensions bring us to our proposed model called \textbf{AutoSense}. The graphical representation of AutoSense is shown in Figure \ref{fig:condsptm}, while the meaning of the notations used in this paper is shown in Table \ref{tab:notations}.

\paragraph{Generative process}

For each instance, we divide the text into two contexts: the local context $L$ which includes the target word $w_t$ and its neighboring words $w_l$, and the global context $M$ which contains the other remaining words $w_m$. Words from different contexts are generated separately.

In the global context $M$, words $w_m$ are generated from either a sense $s$ or a topic $t$ latent variable. The selection is done by a switch variable $x$. If $x=1$, then the word generation is done by using the sense variable $s$. Otherwise, it is done by using the topic variable $t$. The probability of a global context word $w_m$ in document $d$ is given below.
\begin{align*}
    P(w_m|d)
    &= &&P(w_m|x=1) \sum_s P(w_m|s) P(s|d) + \\
    & &&P(w_m|x=2) \sum_t P(w_m|t) P(t|d) \\
    &= &&\theta_{x=1} \sum_s \theta^{(d)}_s \phi^{(s)}_{w_m} + \theta_{x=2} \sum_t \theta^{(d)}_t \phi^{(t)}_{w_m} \nonumber
\end{align*}
In the local context $L$, words $w_l$ are generated from both sense $s$ and topic $t$ variables. Also, the target word $w_t$ is generated along with $w_l$ as target-neighbor pairs $(w_t,w_l)$ using the sense variable $s$. Sense and topic variables are dependent to each other, so we generate them using the joint probability $p(s,t|d)$. We factorize $p(s,t|d)$ approximately using ideas from dependency networks  \cite{heckerman2000dependency} to avoid independency assumptions, i.e. $p(a,b|c)=p(a|b,c)p(b|a,c)$, and deficient modeling \cite{brown1993mathematics} to ignore redundancies, i.e. $p(a|b,c)p(b|a,c)=p(a|b)p(a|c)p(b|a)p(b|c)p(a,b)$. The probability of a local context word $w_l$ in document $d$ given below.

\begin{align*}
    P(w_t,w_l|d)
    &= \sum_s \sum_t && p(w_t|s) p(w_l|s,t) p(s,t|d) \\
    &\approx \sum_s \sum_t && p(w_t|s) p(w_l|s,t) p(s|d,t) p(t|d,s) \\
    &\approx \sum_s \sum_t && p(w_t|s) p(w_l|s) p(w_l|t) \\ & && p(s|d) p(s|t) p(t|d) p(t|s) p(s,t) \\
    &= \sum_s \sum_t && \phi^{(s)}_{w_t} \phi^{(s)}_{w_l} \phi^{(t)}_{w_l} \theta^{(d)}_s \theta^{(d)}_t \theta_{s|t} \theta_{t|s} \theta_{st} \nonumber
\end{align*}
\paragraph{Inference}

We use collapsed Gibbs sampling \cite{griffiths2004finding} to estimate the latent variables. At each transition step of the Markov chain, for each word $w_m$ in the global context, we draw the switch $x\sim \{1,2\}$, and the sense $s=k$ or the topic $t=j$ variables using the conditional probabilities given below. The variable $C^{AB}_{ab}$ represents the number of $a \in A$ and $b \in B$ assignments, excluding the current word. The $rest$ corresponds to the other remaining variables, such as the instance $d$, the current word $w_m$, the $\theta$ and $\phi$ distributions, and the $\alpha$, $\beta$, and $\gamma$ Dirichlet priors.
\begin{multline}
    P(x=1,s=k|rest) = 
    \frac{C^{DX}_{d1} + \gamma}{\sum^2_{x'=1}C^{DX}_{dx'} + 2\gamma}\\
    \frac{C^{DS}_{dk} + \alpha}{\sum^S_{k'=1}C^{DS}_{dk'} + S\alpha}
    \frac{C^{SW}_{kw_m} + \beta}{\sum^{V}_{w'=1}C^{SW}_{kw'} + V_m\beta} \nonumber
\end{multline}
\begin{multline}
    P(x=2,t=j|rest) = 
    \frac{C^{DX}_{d2} + \gamma}{\sum^2_{x'=1}C^{DX}_{dx'} + 2\gamma}\\
    \frac{C^{DT}_{dj} + \alpha}{\sum^T_{j'=1}C^{DT}_{dj'} + T\alpha}
    \frac{C^{TW}_{jw_m} + \beta}{\sum^{V}_{w'=1}C^{TW}_{jw'} + V_m\beta}  \nonumber
\end{multline}
Subsequently, for each word $w_l$ and the target word $w_t$ (forming the target-neighbor pair $(w_t, w_l)$) in the local context, we draw the sense $s=k$ and the topic $t=j$ variables using the conditional probability given below.
\begin{multline}
    P(t_i=j,s_i=k|rest)=
    \frac{C^{DT}_{dj} + \alpha}{\sum^T_{j'=1}C^{DT}_{dj'} + T\alpha}\\
    \frac{C^{DS}_{dk} + \alpha}{\sum^S_{k'=1}C^{DS}_{dk'} + S\alpha}
    \frac{C^{TW}_{jw_l} + \beta}{\sum^{V}_{w'=1}C^{TW}_{jw'} + V_l\beta}\\
    \frac{C^{SW}_{kw_l} + \beta}{\sum^{V}_{w'=1}C^{SW}_{kw'} + V_l\beta}
    \frac{C^{SW}_{kw_t} + \beta}{\sum^{V}_{w'=1}C^{SW}_{kw'} + V_l\beta + 1}\\
    \frac{C^{ST}_{kj} + \alpha}{\sum^T_{j'=1}C^{ST}_{kj'} + T\alpha}
    \frac{C^{TS}_{jk} + \alpha}{\sum^S_{k'=1}C^{TS}_{jk'} + S\alpha}\\
    \frac{C^{ST}_{kj} + \alpha}{\sum^S_{k'=1}\sum^T_{j'=1}C^{ST}_{k'j'} + ST\alpha_{j'}} \nonumber
\end{multline}
\paragraph{Word sense induction}

After inference is done, the approximate probability of the sense $s$ of the target word in a given document $d$ is induced using the sense distribution of the document as shown in the equation below, where $C^{AB}_{ab}$ represents the number of $a \in A$ and $b \in B$ assignments. We also calculate the word distribution of each sense using the second equation below to inspect the meaning of sense.
\begin{equation}
    \label{eq:induction}
    \theta_{s|d} = \frac{C^{DS}_{ds}}{\sum^S_{s'=1}C^{DS}_{ds'}}
    \quad
	\theta_{w|s} = \frac{C^{SW}_{sw}}{\sum^{V}_{w'=1}C^{SW}_{sw'}}
\end{equation}
\section{Experimental setup} 

\paragraph{Datasets and preprocessing}

We use two publicly available datasets: SemEval 2010 Task 14 \cite{manandhar2010semeval} and SemEval 2013 Task 13 \cite{jurgens2013semeval}. The SemEval 2010 dataset\footnote{\url{https://www.cs.york.ac.uk/semeval2010_WSI}} consists of 50 verbs and 50 nouns, each with different number of instances for a total of $8915$ instances. SemEval 2013 dataset\footnote{\url{https://www.cs.york.ac.uk/semeval-2013/task13/}} consists of 20 verbs, 20 nouns, and 10 adjectives, with a total of $4664$ instances.

For preprocessing, we do tokenization, lemmatization, and removing of symbols to build the word lists using Stanford CoreNLP \cite{manning2014stanford}. We divide the word lists into two contexts: the local and global context. Following \cite{wang2015sense}, we set the local context window to 10, with a maximum number of words of 21 (i.e. 10 words before and 10 words after). Other words are put into the global context. Note however that AutoSense has a less strict global/local context assumption as it treats some words in the global context as local depending on the switch variable.

\paragraph{Parameter setting}

We set the hyperparameters to $\alpha=0.1$, $\beta=0.01$, $\gamma=0.3$, following the conventional setup \cite{griffiths2004finding,chemudugunta2006modeling}. We arbitrarily set the number of senses to $S=15$, and the number of topics $T=2S=30$, following \cite{wang2015sense}. We also include four other versions of our model: 
\textbf{AutoSense${}^{-wp}$} removes the target-neighbor pair constraint and transforms the local context to that of STM,
\textbf{AutoSense${}^{-sw}$} removes the switch variable and transforms the global context to that of LDA,
\textbf{AutoSense${}^{s=X}$} is a tuned and best version of the model, where the number of senses is tuned over a separate development set provided by the shared tasks and $X$ is the tuned number of sense, different for each dataset, and
\textbf{AutoSense${}^{s=100}$} is the overestimated and worst version of the model, where we set the number of senses to an arbitrary large number, i.e. 100.

We set the number of iterations to $2000$ and run the Gibbs sampler. Following the convention of previous works \cite{lau2012word,goyal2014unsupervised,wang2015sense}, we assume convergence when the number of iterations is high. However, due to the randomized nature of Gibbs sampling, we report the average scores over 5 runs of Gibbs sampling. We then use the distribution $\theta_{s|d}$ as shown in Equation \ref{eq:induction} as the solution of the WSI problem.

\section{Experiments} 

\subsection{Word sense induction} 
\label{sec:results}

\paragraph{SemEval 2010} 

For the SemEval 2010 dataset, we compare models using two unsupervised metrics: V-measure (\textbf{V-M}) and paired F-score (\textbf{F-S}). V-M favors a high number of senses (e.g. assigning one cluster per instance), while F-S favors a small number of senses (e.g. all instances in one cluster) \cite{manandhar2010semeval}. In order to get a common ground for comparison, we do a geometric average \textsc{AVG} of both metrics, following \cite{wang2015sense}. Finally, we also report the absolute difference between the actual (3.85) and induced number of senses as $\delta(\#S)$.

We compare with seven other models:
a) LDA on co-occurrence graphs (\textbf{LDA}) and 
b) spectral clustering on co-occurrence graphs (\textbf{Spectral}) as reported in  \cite{goyal2014unsupervised},
c) Hidden Concept (\textbf{HC}),
d) HC using Zipf's law (\textbf{HC+Zipf}), and 
e) Bayesian nonparametric version of HC (\textbf{BNP-HC}) as reported in  \cite{chang2014inducing},
f) CRP-based sense embeddings with positive PMI vectors as pre-trained vectors (\textbf{CRP-PPMI}), and 
g) Multi-Sense Skip-gram Model (\textbf{SE-WSI-fix}) as reported in \cite{song2016word}.

\begin{table}[t]
    \small
	\centering
	\begin{subtable}{0.47\textwidth}
		\centering
        \begin{tabular}{|c|c|cc|c|c|}
        \hline
        \multicolumn{2}{|c|}{Model} & F-S   & V-M   & \textsc{AVG}   & $\delta(\#S)$ \\
        \hline
        \multicolumn{1}{|c|}{\multirow{5}[2]{*}{\makecell{LVMs}}} & LDA   & 60.7  & 4.4   & 16.34 & 1.40 \\
              & Spectral & 61.5  & 4.5   & 16.64 & 1.98 \\
              & HC    & 44.4  & 11.5  & 22.62 & 1.15 \\
              & HC+Zipf & 35.1  & 15.2  & 23.10 & 3.81 \\
              & BNP-HC & 23.1  & \textbf{21.4} & 22.23 & 11.77 \\
        \hline
        \multicolumn{1}{|c|}{\multirow{2}[2]{*}{\makecell{NBEs}}} & CRP-PPMI & 57.7  & 2.9   & 12.94 & 2.09 \\
              & SE-WSI-fix & 55.1  & 9.8   & 23.24 & 1.35 \\
        \hline
        \multicolumn{2}{|c|}{AutoSense$^{-wp}$} & 59.3  & 9.2   & 23.36 & 2.16 \\
        \multicolumn{2}{|c|}{AutoSense$^{-sw}$} & 61.1  & 8.6   & 22.92 & 1.42 \\
        \multicolumn{2}{|c|}{AutoSense}   & 61.7  & 9.8   & 24.59 & 0.33 \\
        \multicolumn{2}{|c|}{AutoSense$^{s=5}$} & \textbf{62.9} & 10.1  & \textbf{25.20} & \textbf{0.32} \\
        \multicolumn{2}{|c|}{AutoSense$^{s=100}$} & 61.2  & 9.6   & 24.23 & 0.78 \\
        \hline
        \end{tabular}%
		\caption{SemEval 2010 WSI dataset}    
		\label{tab:semeval2010}
	\end{subtable}
	\newline
    \vspace*{2pt}
    \newline
	\begin{subtable}{0.47\textwidth}
		\centering
        \begin{tabular}{|c|c|cc|c|}
        \hline
        \multicolumn{2}{|c|}{Model} & F-BC  & F-NMI & \textsc{AVG} \\
        \hline
        Substitution & AI-KU & 39.0  & 6.50  & 15.92 \\
        \hline
        \multicolumn{1}{|c|}{\multirow{2}[2]{*}{\makecell{LVMs}}} & Unimelb & 48.3  & 6.00  & 17.02 \\
              & STM   & 53.5  & 6.96  & 19.30 \\
        \hline
        \multicolumn{1}{|c|}{\multirow{2}[2]{*}{\makecell{NBEs}}} & WG    & 58.1  & 1.60  & 9.64 \\
              & DIVE  & 49.9  & 3.50  & 13.22 \\
        \hline
        \multirow{2}[2]{*}{LVMs + NBEs} & STM+w2v & 55.4  & 7.14  & 19.89 \\
              & MCC   & 55.6  & 7.62  & 20.58 \\
        \hline
        \multicolumn{2}{|c|}{AutoSense$^{-wp}$} & 55.7  & 7.69  & 20.69 \\
        \multicolumn{2}{|c|}{AutoSense$^{-sw}$} & 61.4  & 7.36  & 21.26 \\
        \multicolumn{2}{|c|}{AutoSense}   & \textbf{61.7} & 7.96  & 22.16 \\
        \multicolumn{2}{|c|}{AutoSense$^{s=7}$} & \textbf{61.7} & \textbf{7.97} & \textbf{22.17} \\
        \multicolumn{2}{|c|}{AutoSense$^{s=100}$} & 61.0  & 7.25  & 21.03 \\
        \hline
        \multicolumn{5}{|c|}{\textit{with additional contexts}} \\
        \hline
        \multirow{2}[2]{*}{STM} & +actual & {59.1}  & 9.39  & {23.56} \\
              & +ukWac & 54.5  & \textbf{9.74}  & 23.04 \\
        \hline
        AutoSense & +actual & \textbf{62.2} & {9.55} & \textbf{24.37} \\
        \hline
        \end{tabular}%
		\caption{SemEval 2013 WSI dataset}
		\label{tab:semeval2013}
	\end{subtable}
	\caption{Performance of different models on the datasets. Best scores are bold-faced. LVMs are Latent Variable Models, while NBEs are Neural-based Embeddings.}
	\label{tab:results}
\end{table}

Results are shown in Table \ref{tab:semeval2010}, where AutoSense outperforms other competing models on \textsc{AVG}. Among the AutoSense models, the AutoSense${}^{-wp}$ and AutoSense${}^{-sw}$ version perform the worst, emphasizing the necessity of the target-neighbor pairs and the switch variable. The overestimated AutoSense${}^{s=100}$ performs better than previously proposed models, proving the robustness of our model on the different word sense granularities. On the $\delta(\#S)$ metric, the untuned AutoSense and AutoSense${}^{s=5}$ perform the best. The V-M metric needs to be interpreted carefully, because it can easily be maximized by separating all instances into different sense clusters and thus overestimating the actual number of senses $\#S$ and decreasing the F-S metric. The model BNP-HC is an example of such: Though its V-M metric is the highest, it scores the lowest on the F-S metric and greatly overestimates $\#S$, thus having a very high $\delta(\#S)$. The goal is thus a good balance of V-M and F-S (i.e. highest \textsc{AVG}), and a close estimation of $\#S$ (i.e. lowest $\delta(\#S)$, which is successfully achieved by our models.

\paragraph{SemEval 2013} 

Two metrics are used for the SemEval 2013 dataset: fuzzy B-cubed (\textbf{F-BC}) and fuzzy normalized mutual information (\textbf{F-NMI}). F-BC gives preference to labelling all instances with the same sense, while F-NMI gives preference to labelling all instances with distinct senses. Therefore, computing the \textsc{AVG} of both metrics is also necessary in this experiment, for ease of comparison, as also suggested in \cite{wang2015sense}.

We use seven baselines:
a) lexical substitution method (\textbf{AI-KU}) and
b) nonparametric HDP model (\textbf{Unimelb}) as reported in \cite{jurgens2013semeval},
c) Sense-Topic Model \textbf{STM},
d) STM with word2vec weights (\textbf{STM+w2v}) as reported in  \cite{wang2015sense},
e) Word Graph embeddings (\textbf{WG}), 
f) Distributional Inclusion Vector Embedding (\textbf{DIVE}) as reported in \cite{chang2018efficient},
and
g) Multi Context Continuous model \textbf{MCC} as reported in \cite{komninos2016structured}. 

Results are shown in Table \ref{tab:semeval2013}. Among the models, all versions of AutoSense perform better than other models on \textsc{AVG}. The untuned AutoSense and AutoSense$^{s=7}$ especially garner noticeable increase of $6.1\%$ on fuzzy B-cubed metric from MCC, the previous best model. We also notice a big $6.0\%$ decrease on the fuzzy B-cubed of AutoSense when the target-neighbor pair context is removed. This means that introducing the target-neighbor pair is crucial to the improvement of the model. Finally, the overestimated AutoSense model performs as well as the other AutoSense models, even outperforming all previous models on \textsc{AVG}, which proves the effectiveness of AutoSense even when $s$ is set to a large value.

For completeness, we also report STM with additional contexts, STM+actual and STM+ukWac \cite{wang2015sense}, where they used the actual additional contexts from the original data and semantically similar contexts from ukWac, respectively, as additional global context. With the performance gain we achieved, AutoSense without additional context can perform comparably to models with additional contexts: Our model greatly outperforms these models on the F-BC metric by at least 2\%. Also, considering that both AutoSense and STM are LDA-based models, the same data enhancements can straightforwardly be applied when the needs arise. We similarly apply the actual additional contexts to AutoSense and find that we achieve state-of-the-art performance on \textsc{AVG}.

\subsection{Sense granularity problem} 
\label{sec:sensegran}

The main weakness of LDA when used on WSI task is the sense granularity problem. Recent models such as HC \cite{chang2014inducing} and STM \cite{wang2015sense} mitigated this problem by tuning the number of senses hyperparameter $S$ to minimize the error. However, such tuning, often empirically set to a small number such as $S=3$ \cite{wang2015sense}, fails to infer varying number of senses of words, especially for words with a higher number of senses. Nonparametric models such as HDP and BNP-HC \cite{lau2013unimelb,chang2014inducing} claim to automatically induce different $S$ for each word. However, as shown in the results in Table \ref{tab:results}, the estimated $S$ is far from the actual number of senses and both models are ineffective.

\begin{table}[!t]
	\centering
	\begin{tabular}{|c|l|r|}
		\hline
		Sense & Word distribution & \#Docs \\ \hline
		1 & hotel tour tourist summer flight & 22 \\ \hline
		2 & month ticket available performance & 3 \\ \hline
		3 & guest office stateroom class suite & 3 \\ \hline
		* & advance overseas line popular japan & 0 \\ \hline
		* & email day buy unable tour & 0 \\ \hline
		* & sort basic tour time & 0 \\ \hline
	\end{tabular}
	\caption{Six of the 15 senses of the target verb \textit{book} using AutoSense with $S=15$.
		The word lists shown are preprocessed to remove stopwords and the target word.
		The first three senses are senses which are assigned at least once to an instance document. The last three are \textit{garbage senses}.}
	\label{tab:sample}
\end{table}
\label{sec:senses}

On the other hand, Table \ref{tab:results} also shows that AutoSense is effective even when $S$ is overestimated. We explain why through an example result shown in Table \ref{tab:sample}, where the target word is the verb \textit{book}, the actual number of senses is three, and $S$ is set to 15. First, we see that there are senses which are not assigned to any instance document, signified by $*$, which we call \textbf{garbage senses}. We notice that effectively representing a new latent variable for sense as a distribution over topics forces the model to throw garbage senses. Second, while it is easy to distinguish the third sense (i.e., \textit{book: register in a booker}) to the two other senses, the first and second senses both refer to planning or arranging for an event in advance. Incorporating the target-neighbor pairs helps the model differentiates both into \textbf{fine-grained senses} \textit{book: arrange for and reserve in advance} and \textit{book: engage for a performance}.

\begin{figure}[!t]
	\centering
	\begin{subfigure}{0.45\textwidth}
	    \centering
		\includegraphics[width=\textwidth]{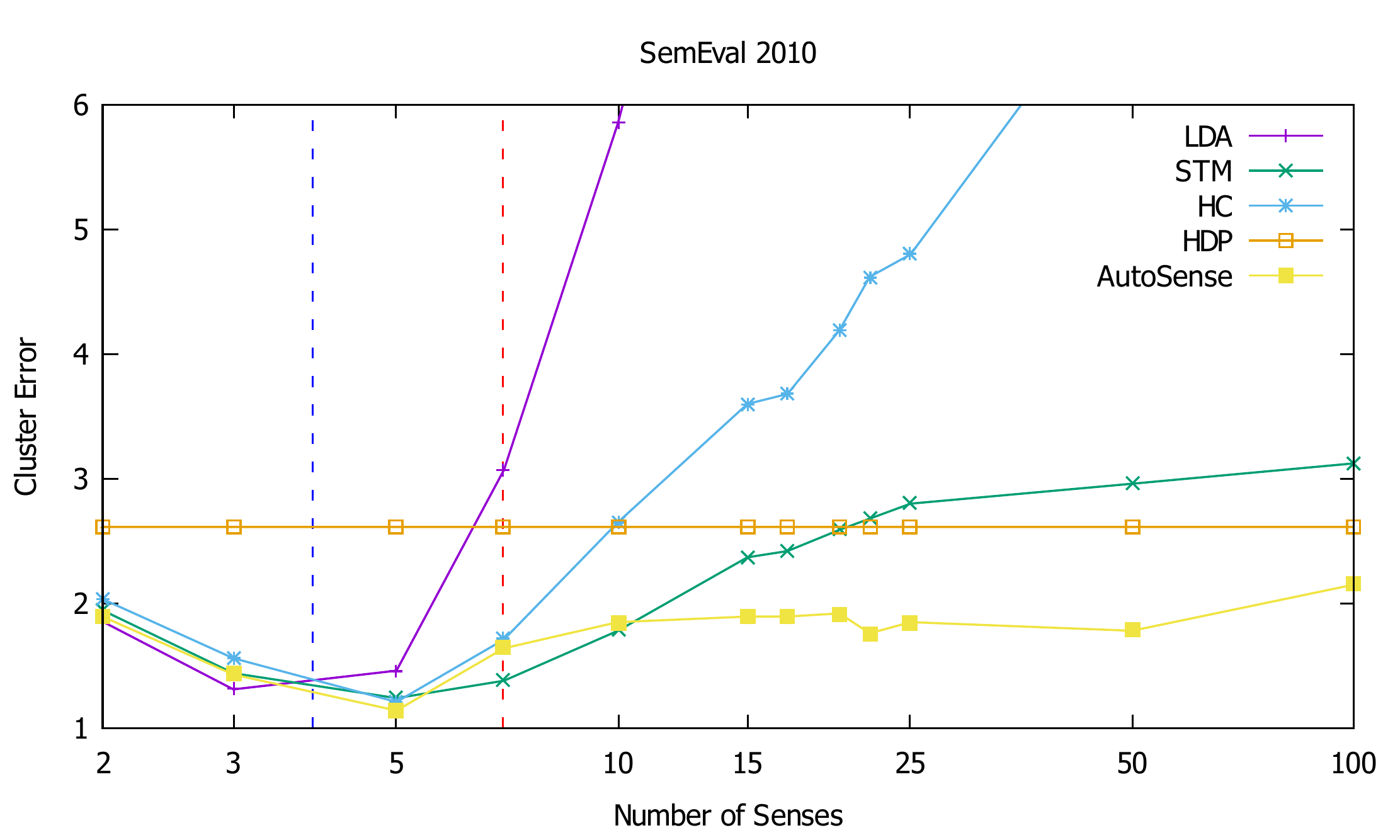}
		\caption{SemEval 2010 dataset}
	\end{subfigure}
	\begin{subfigure}{0.45\textwidth}
	    \centering
		\includegraphics[width=\textwidth]{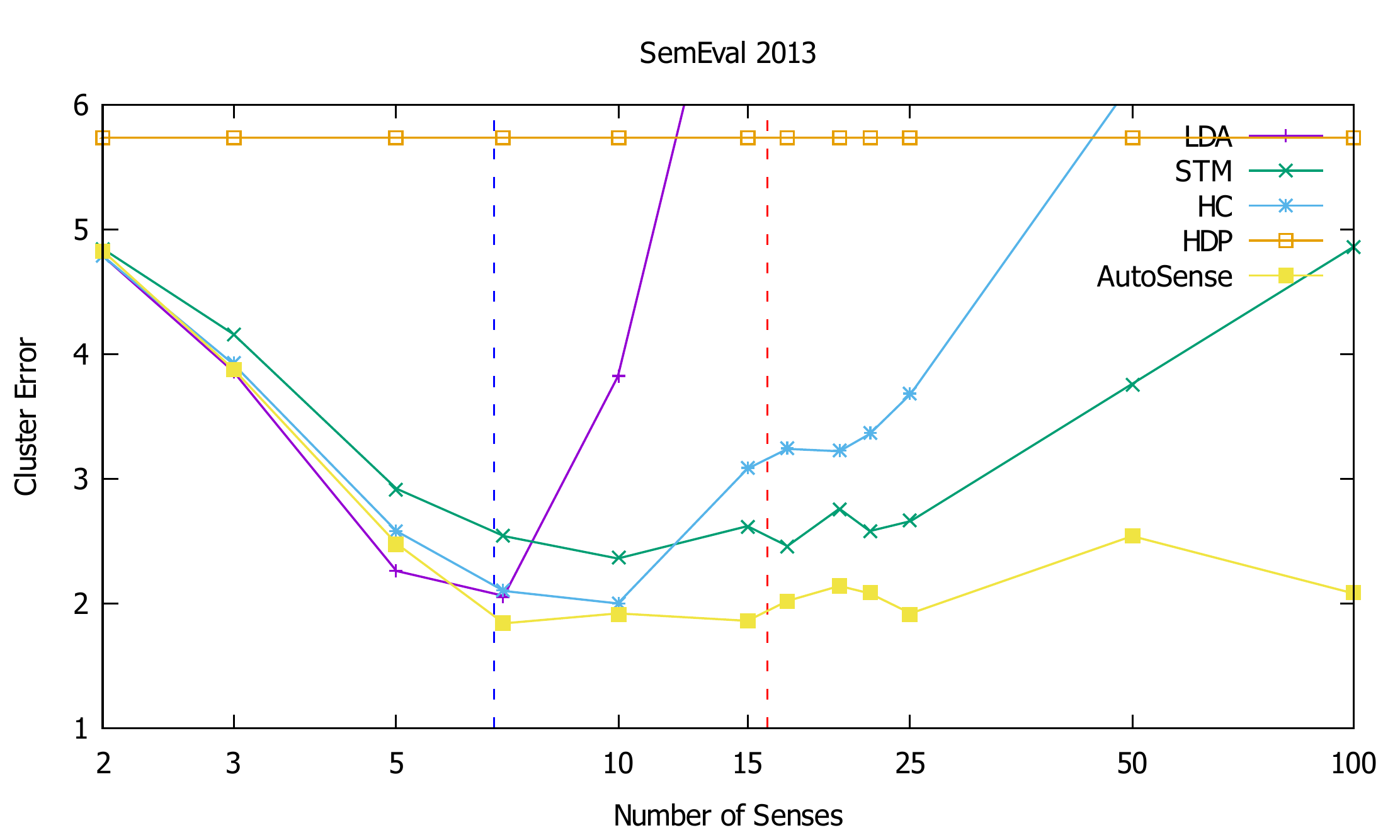}
		\caption{SemEval 2013 dataset}
	\end{subfigure}
	\caption{Cluster error of models with different number of senses $S$. The vertical dashed lines correspond to the \textcolor{blue}{mean} and the \textcolor{red}{max} of the actual number of senses. The x-axes are log-scaled.}
	\label{fig:clusters}
\end{figure}

We compare the competing models quantitatively on how they correctly detect the actual number of sense clusters using \textbf{cluster error}, which is the mean absolute error between the detected number and the actual number of sense clusters. We compare the cluster errors of LDA \cite{blei2003latent}, STM \cite{wang2015sense}, HC \cite{chang2014inducing}, and a nonparametric model HDP \cite{teh2004sharing}, with AutoSense. We report the results in Figure \ref{fig:clusters}. Results show that the cluster error of LDA increases sharply as the number of senses exceeds the actual mean number of senses. HC and STM also throw garbage senses since they also introduce in some way a new sense variable, however the cluster errors of both models still increase when $S$ is set beyond the maximum number of senses. We argue that this is because first, the sense representation is not optimal as they assume strict local/global context assumption, and second and most importantly, the models do not produce fine-grained senses. AutoSense does both garbage sense throwing and fine-grained sense induction, which helps in the detection of the actual word granularity. Finally, the cluster error of AutoSense is always better than that of HDP. This shows that AutoSense, despite being a parametric model, automatically detects the number of sense clusters without parameter tuning and is more accurate than the automatic detection of nonparametric models.

\subsection{Unsupervised author name disambiguation}

Unsupervised author name disambiguation (UAND) is a task very similar to the WSI task, where ambiguous author names are the target words. However, one additional challenge of UAND is that there can be as many as 100 authors with the same name, whereas words can have at most 20 different senses, at least in our datasets, as shown in the dataset statistics in Table \ref{tab:stat}. Moreover, the standard deviations of the author name disambiguation datasets are also higher, which means that there is more variation on the number of senses per target author name. Thus, in this task, the sense granularity problem is more difficult and needs to be addressed properly.

\begin{table}[t]
    \centering
    \begin{tabular}{|c|cccc|}
        \hline
        Dataset & Min & Max & Mean & StdDev \\
        \hline
        SemEval 2010 & 2 & 16 & 7.68 & 3.35 \\ \hline
        SemEval 2013 & 2 & 7 & 3.85 & 1.40 \\ \hline
        PubMed & 1 & 28 & 10.41 & 7.68 \\ \hline
        Arnet & 1 & 112 & 14.18 & 18.02 \\
        \hline
    \end{tabular}
    \caption{Statistics of the number of senses of target words/names in the datasets used in the paper.}
    \label{tab:stat}
\end{table}

Current state-of-the-art models use non-text features such as publication venue and citations \cite{tang2012unified}.  We argue that text features also provide informative clues to disambiguate author names. In this experiment, we make use of text features such as the title and abstract of research papers as data instance of the task. In addition, we also include in our dataset author names and the publication venue as pseudo-words. In this way, we can reformulate the UAND task as a WSI task, and exploit text features not used in current techniques.

\paragraph{Experimental setup}

We use two publicly available datasets for the UAND task: Arnet\footnote{\url{https://aminer.org/disambiguation}} and PubMed\footnote{\url{https://github.com/Yonsei-TSMM/author_name_disambiguation}}. The Arnet dataset contains 100 ambiguous author names and a total of 7528 papers as data instance. Each instance includes the title, author list, and publication venue of a research paper authored by the given author name. In addition, we also manually extract the abstracts of the research papers for additional context.
The PubMed dataset contains 37 author names with a total of 2875 research papers as instances. It includes the PubMed ID of the papers authored by the given author name. We extract the title, author list, publication venue, and abstract of each PubMed ID from the PubMed website.

We use LDA \cite{blei2003latent}, HC \cite{chang2014inducing} and STM \cite{wang2015sense} as baselines. We do not compare with non-text feature-based models \cite{tang2012unified,cen2013author} because our goal is to compare sense topic models on a task where the sense granularities are more varied. For STM and AutoSense, the title, publication venue and the author names are used as local contexts while the abstract is used as the global context. This decision is based on conclusions from previous works \cite{tang2012unified} that the title, publication venue, and the author names are more informative than the abstract when disambiguating author names. We use the same parameters as used above, and we set $S$ to 5, 25, 50, and 100 to compare the performances of the models as the number of senses increases.

\begin{table}[]
    \centering
    \begin{subtable}{0.47\textwidth}
    \centering
    \begin{tabular}{|c|cccc|}
        \hline
        Model & $S=5$ & $S=25$ & $S=50$ & $S=100$ \\
        \hline
        LDA & 31.5 & 13.4 & 9.8 & 8.2 \\
        HC & 46.3 & 46.3 & 44.4 & 41.7 \\
        STM & 52.8 & 55.0 & 55.5 & 55.0 \\
        AutoSense & 56.2 & 56.4 & 57.9 & 58.8 \\
        \hline
    \end{tabular}
    \caption{Arnet Dataset}
    \end{subtable}
	\newline
    \vspace*{2pt}
    \newline
    \begin{subtable}{0.47\textwidth}
    \centering
    \begin{tabular}{|c|cccc|}
        \hline
        Model & $S=5$ & $S=25$ & $S=50$ & $S=100$ \\
        \hline
        LDA & 41.4 & 13.3 & 8.9 & 9.0 \\
        HC & 42.5 & 44.1 & 41.6 & 41.3 \\
        STM & 44.9 & 44.4 & 44.9 & 41.9 \\
        AutoSense & 44.4 & 45.5 & 46.6 & 46.5 \\
        \hline
    \end{tabular}
    \caption{PubMed Dataset}
    \end{subtable}
    \caption{Paired F1 measures of competing models with different number of senses $S$ on UAND datasets.}
	\label{fig:andresults}
\end{table}

\paragraph{Results}

For evaluation, we use the pairwise F1 measure to compare the performance of competing models, following \cite{tang2012unified}. Results are shown in Figure~\ref{fig:andresults}. AutoSense performs the best on almost all settings, except on the PubMed dataset and when $S=5$, where it garners a comparable result with STM. However, in the case where $S$ is set close to the maximum number of senses in the dataset (i.e. 28 in PubMed and 112 in Arnet), AutoSense performs the best among the models. LDA and HC perform badly on all settings and greatly decrease their performances when $S$ becomes high. STM also shows decrease in performance on the PubMed dataset when $S=100$. This is because the PubMed dataset has a lower maximum number of senses, and STM is sensitive in the setting of $S$, and thus hurts the robustness of the model to different sense granularities. 

\section{Conclusion} 

We proposed a solution to answer the sense granularity problem, one of the major challenges of the WSI task. We introduced AutoSense, a latent variable model that not only throws away garbage senses, but also induces fine-grained senses. We showed that AutoSense greatly outperforms the current state-of-the-art models in both SemEval 2010 and 2013 WSI datasets. We also show experiments on how AutoSense is able to overcome sense granularity problem, a well-known flaw of latent variable models on. We further applied our model to UAND task, a similar task but with more varying number of senses, and showed that AutoSense performs the best among latent variable models, proving its robustness to different sense granularities.

\section*{Acknowledgements}

This work was supported by Samsung Research Funding
Center of Samsung Electronics under Project Number SRFCIT1701-01, and by Next-Generation Information Computing Development Program through the National Research Foundation of Korea (NRF), funded by the Ministry of Science, ICT (No.NRF-2017M3C4A7065887).

\bibliography{aaai19}
\bibliographystyle{aaai}

\end{document}